\documentclass{article}
\pdfoutput=1
\usepackage{flushend}
\usepackage{spconf,amsmath,graphicx}
\usepackage{combelow}
\usepackage{algorithm}
\usepackage{algorithmic}
\usepackage{hyperref}
\title{REFINING THE BOUNDING VOLUMES FOR LOSSLESS COMPRESSION OF VOXELIZED POINT CLOUDS GEOMETRY}
 \name{Emre Can Kaya$^{\star}$ \qquad Sebastian Schwarz$^{\dagger}$ \qquad Ioan Tabus$^{\star}$}

\address{$^{\star}$ Tampere University, Tampere, Finland \\
    $^{\dagger}$ Nokia Technologies, Munich, Germany}

\begin{document}

\maketitle
\begin{abstract}
This paper describes a novel lossless compression method for point cloud geometry, building on a recent lossy compression method that aimed at reconstructing only the bounding volume of a point cloud.  The proposed scheme starts by partially reconstructing the geometry from the two depthmaps associated to a single projection direction. The partial reconstruction obtained from the depthmaps is  completed to a full reconstruction of the point cloud by sweeping section by section along one direction and encoding the points which were not contained in the two depthmaps. The main ingredient is a list-based encoding of the inner points (situated inside the feasible regions) by a novel arithmetic three dimensional context coding procedure that efficiently utilizes rotational invariances present in the input data. State-of-the-art bits-per-voxel results are obtained on benchmark datasets. 
\end{abstract}

\section{Introduction}
\label{sec:intro}
The lossless compression of point clouds was thoroughly studied and is currently under standardization under MPEG  \cite{schwarz2018emerging},\cite{mpeg} and JPEG activities \cite{ebrahimi2016jpeg}. The research literature on point cloud compression includes a lot of methods based on octree representations, e.g., \cite{raht,mekuria2016design,milani2017fast,garcia2018intra,garcia2019geometry}. The current GPCC lossless method \cite{gpcc} is also based on octree representations \cite{octree}, where a point cloud is represented as an octree, which can be parsed from the root to the leaves and at each depth level in the tree, one obtains a lossy reconstruction of the point cloud at a certain resolution, while the lossless reconstruction is obtained at the final resolution level in the octree. At each resolution level, an octree node corresponds to a cube at a particular 3D location and the octree node is labeled by one if within that cube there is at least one final point of the point cloud. By traversing in a breadth-first way, one can obtain a lossy reconstruction at each depth of the tree, while the lossless reconstruction is obtained at the final resolution. Octree-based approach is attractive for providing a progressive-in-resolution reconstruction. %On the other hand it has the disadvantage that the lossless reconstruction is obtained by necessarily performing all encodings-decodings at each depth level of the tree. Very well engineered solutions were provided in GPCC for conditional coding of each bit of the occupancy pattern at each node of the octree, exploiting the neighbors of the nodes in the octree by acquiring the needed context from the parent node, and its 3D neighbor nodes, or from the already encoded children of the parent’s neighbors.

Recently, good lossless performance was achieved by methods based on successive decomposition of the point cloud data into binary images \cite{bdc,dyadic}, without using octree representations.

This paper describes a novel lossless point cloud compression algorithm which is based on Bounding Volumes \cite{bv} for G-PCC. In the Bounding Volumes, all reconstructed points truly belonged to the point cloud, but some points, specifically inner points in the transversal sections of the point cloud, were not encoded at all. In this work, a complete lossless process, overlapping in the first stage with the Bounding Volumes method (encoding a front and a back projection), but diverging from the previous method, already at the second stage, that of encoding the true boundary of the feasible region and making it less restrictive, getting rid of the requirement of decomposing the point cloud into tubes having single connected component sections. 

Compared to the lossy Bounding Volume method, the newly introduced encoding process includes an extensive context coding scheme that can finally provide a lossless reconstruction of the point cloud. This novel scheme is based on carefully selected context coding methods and intensively uses exclusion to avoid unnecessary testing for the occupancy of the locations which are already known.

\section{Proposed Method}
\label{sec:Proposed}

\begin{figure}[tb]

  \centering
  \centerline{\includegraphics[width=\linewidth]{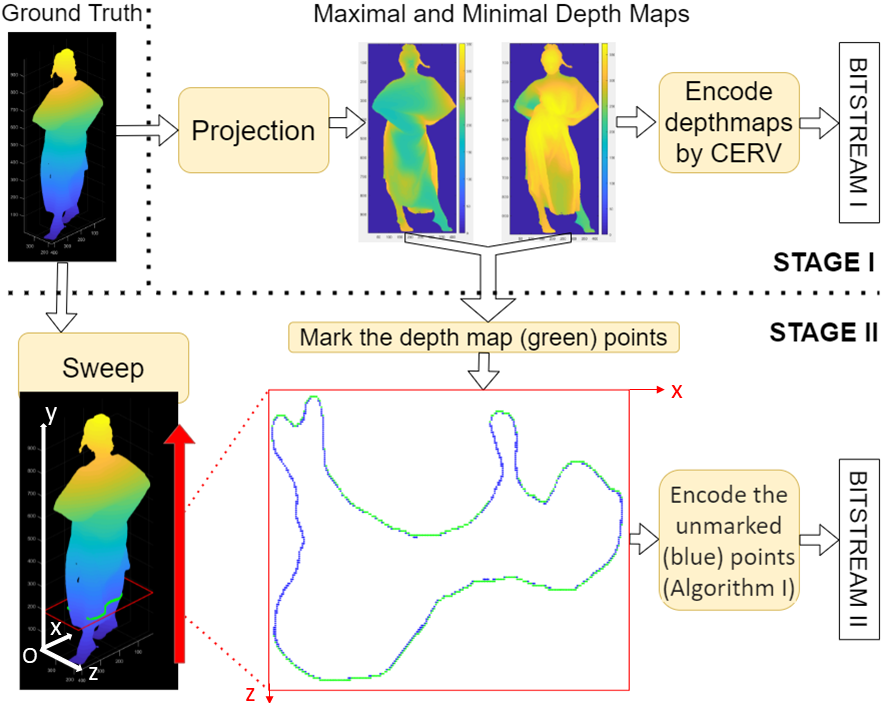}}
%  \vspace{2.0cm}
  \centerline{\parbox{\linewidth}{} }\medskip

\caption{Overview of the proposed method: In Stage I, minimal and maximal depthmaps $Z_{min}$ and $Z_{max}$ are generated by projecting the point cloud along z axis. The depthmaps are encoded by CERV\cite{cerv}. In Stage II, the point cloud is swept through y axis. The points that are not already encoded by CERV (blue points) are encoded as described in Sections \ref{sec:stage2} and \ref{sec:normcontx}.}
\label{fig:algo}
\end{figure}

Consider a point cloud having the resolutions $N_x$, $N_y$, $N_z$ along the axes $x$, $y$, $z$, respectively. The points are encoded in two stages. In Stage I, two projections of the points cloud are encoded; the front and the back projections on xy plane. These projections are two depthmap images, each with $N_x$x$N_y$ pixels. Then, the coding proceeds along the Oy axis of the 3D coordinate system, drawing transverse sections of the point cloud parallel to the zOx plane and encoding each such section in Stage II. An overview of the method is presented on Fig. \ref{fig:algo}. The regularity of the geometric shapes, including smoothness of the surface and the continuity of the edges, are exploited by using context models, where the probability of occupancy of a location is determined by the occupancy of its neighbor locations in 3D space, by including the causal (previously encoded) 3D neighbors from the current and the past sections. 
%During encoding, efficient context models relying on a model of watertight outer surface of the point cloud are utilized. We utilize such a primitive in the encoding process, which leads to exclusion of sets of non-occupied points as large as possible within the current section. 
\subsection{Stage I: Encoding a front and a back depthmap projection }
\label{sec:stage1}

The first encoding stage is intended for defining and  encoding the maximal and minimal depthmaps (representing heights above the Oxy plane) resulting in exclusion sets containing large parts of the space. The minimal depthmap, $Z_{min}$, has at the pixel $(x,y)$ the value, $Z_{min}(x,y)$ equal to the minimum z for which $(x, y, z)$ is a point in the original point cloud. Similarly, the maximal depthmap, $Z_{max}$ has at the pixel $(x,y)$ the value, $Z_{max}(x,y)$ equal to the maximum z for which $(x, y, z)$ is a point in the original point cloud. If no point with $(x,y)$ exists in the point cloud, then it is set $Z_{min}(x,y) = Z_{max}(x,y) = 0$. The encoding of these depthmaps is performed by CERV \cite{cerv}, which encodes the geodesic lines using contexts very rich in information.

\subsection{Stage II: Encoding the remaining points  }
\label{sec:stage2}

At Stage II, we sweep the point cloud along the y dimension, stepping $y_0$ from 0 to $N_y-1$, and we reconstruct all the points in the current section $y_0$ in the ($N_z$x$N_x$) binary image $R$ (current reconstruction). At every $y_0$, points projected to the depthmaps are already known to the decoder and we initialize the current reconstruction $R$ with the projected points such that, $R(Z_{max}(x,y_0),x)=1$ and $R(Z_{min}(x,y_0),x)=1$ for every  $Z_{max}(x,y_0)>0$ and $Z_{min}(x,y_0)>0$. $R$ will be updated whenever a new point is encoded or decoded. We note that, in the binary image $R$, we know at each column x which are the lowest and the highest occupied points (minimal and maximal depths). We consequently construct the binary image $F$ of feasible regions, i.e., of locations in the plane that are possibly occupied (the magenta pixels in Fig. \ref{fig:2}(b)). Formally, $F(z,x) = 1$ for all $(z,x)$ pair satisfying $Z_{min}(x,y_0) \le z \le Z_{max}(x,y_0)$. Using $R$ and $F$, we initialize a binary image $K$, where 1 denotes that the occupancy of a location is known. For example, the locations outside the feasible region are known to be unoccupied, hence, $K(z,x)=1$ for those locations. The true points in the section, that we need to losslessly encode, are marked in a binary image denoted $T$ (see Fig. \ref{fig:2}(a)), and the reconstructed points in the past section (at $y_0-1$, that is already known to the decoder) are marked in a binary image $P$. 

In the image R the already reconstructed true points belonging to depthmaps form a set $\phi$ of pixels. We perform a morphological dilation of the set $\phi$ using as structural element the $3\times3$ element. This obtained set of locations is traversed along the rows of the 2D image and is stored in a list $L$. We also initialize a binary marker image $M$ to mark the pixels already in the list $L$. After this initialization step, the list $L$ is processed sequentially, starting from the top of the list, processing a current pixel $(z, x)$. Both encoder and decoder check whether $K(z, x) = 1$, and if yes, the point is removed from the list, since its occupancy status is already known. Otherwise, if $K(z, x) = 0$, we transmit the value of $T(z, x)$ using arithmetic coding with the coding distribution stored at the context $\zeta$. After that, the counts of symbols associated with the context $\zeta$ are updated. $K$ is updated as $K(z, x) \leftarrow 1$ , and the reconstructed image is updated as $R(z, x) \leftarrow T(z, x)$. If the value $T(z, x) = 1$, we include to the list any neighbor $(z_n, x_n )$ of $(z, x)$, (in 8-connectivity) for which $K(z_n, x_n ) = 0$ and for which the marked status is 0, $M(z_n, x_n) = 0$. After inclusion, the marked status is set to 1, $M(z_n, x_n ) = 1$. This procedure is repeated until the list becomes empty. At the end, we have encoded all the points that are connected to the boundary of the feasible region by a path of connected pixels (in 8-connectivity). After the final section $y_0 = N_y-1$ is processed, all the points that are connected to the points contained in the initial two depthmap images are encoded. For the voxelized point clouds, this outer shell of points contains the majority of the points in the point cloud. The remaining points (if any) are encoded by processing additional shells, as described in Section \ref{sec:peeloff}.

\subsection{Normalized Contexts }
\label{sec:normcontx}

\begin{figure}[ht!]

  \includegraphics[width=0.49\linewidth]{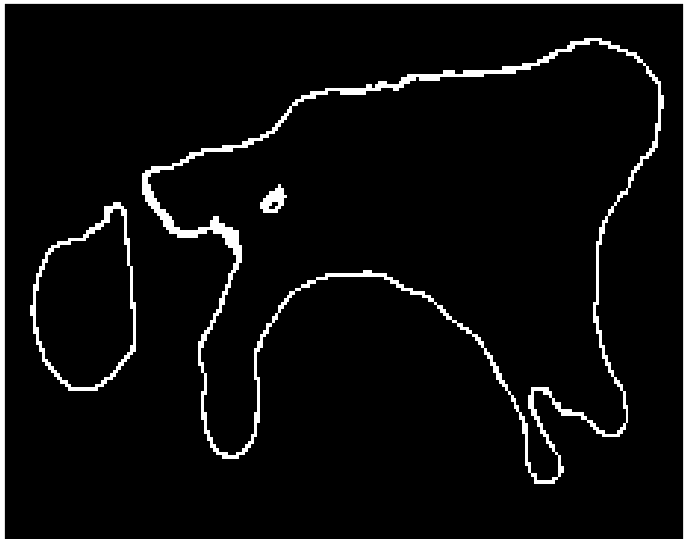}
\includegraphics[width=0.49\linewidth]{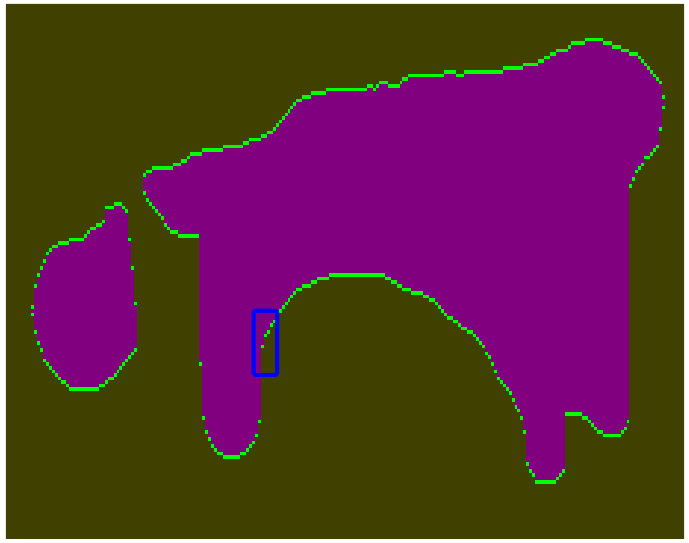}
\centerline{(a)\hspace{3.5cm}(b)}

\vspace{0.5cm}
 \includegraphics[width=0.49\linewidth]{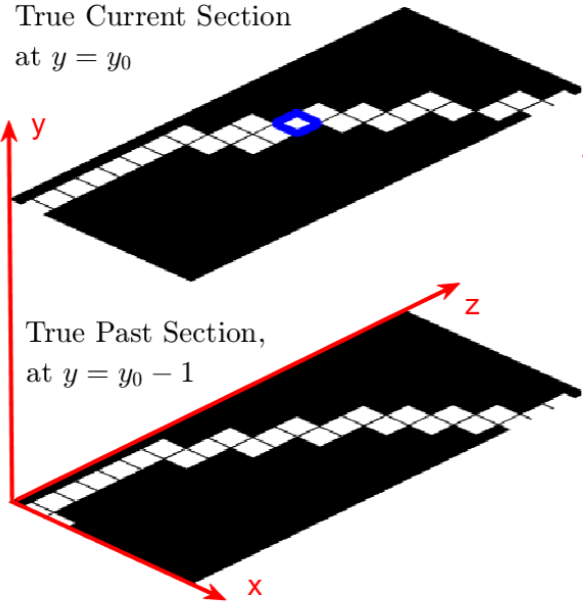}
 \includegraphics[width=0.49\linewidth]{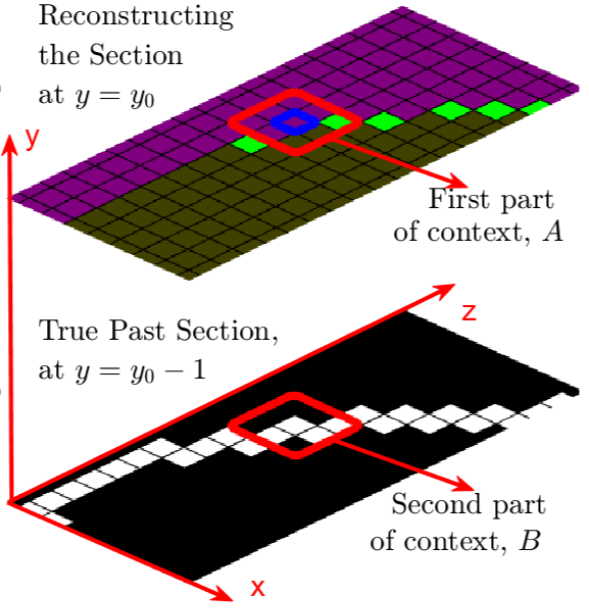}
 \centerline{(c)\hspace{3.5cm}(d)}
%  \vspace{2.0cm}

\caption{(a) True current section; (b) The feasible region (magenta), points from the depth maps (green), forbidden region (khaki); (c) Details of the ground truth inside the blue rectangle from (b); (d) Selection of the two part context for encoding the blue marked pixel}
\label{fig:2}
\end{figure}

\begin{algorithm}[!ht]
\label{algo1}
\caption{Stage II of encoding}
\begin{algorithmic} 

\REQUIRE 

$T$: True section binary image at $y=y_0$ ($N_z$ x $N_x$)

$P$: True section binary image at $y=y_0-1$ ($N_z$ x $N_x$)

$R$: Current reconstruction at $y=y_0-1$ ($N_z$ x $N_x$)

$F$: Feasible Regions bin. image derived from $R$ (Sect. \ref{sec:stage2})

$K$: Binary image of known locations $K \leftarrow  \overline{F} + R$ 

%$K$:Binary image where a location is set to 1 if it is already known whether it is occupied or not ($N_z$ x $N_x$)

%\STATE $R \leftarrow Pr$ ($R$: Current Reconstruction)
$L$: Pixels to be processed $L \leftarrow  \{(z,x) \ni R(z,x)=1\}$
%\ENSURE $y = x^n$

$M$: Binary image of pixels that has been to $L$. $M \leftarrow R$

\WHILE{$L \neq \emptyset$}
\STATE Read $(z,x)$ from the top of $L$
\STATE Extract a 3x3 matrix $R_{3x3}$ from $R$ centered at $(z,x)$
\STATE Extract a 3x3 matrix B from $P$ centered at $(z,x)$
\STATE Extract a 3x3 matrix $K_{3x3}$  by cropping K around   $(z,x)$
\STATE $A \leftarrow R_{3x3} + K_{3x3}$
\STATE Find normalizing rotation $\alpha^{\ast}(A)$ and form $A_{\alpha^{\ast}}$
\STATE Use $\alpha^{\ast}(A)$  to rotate B as $B_{\alpha^{\ast}}$
\STATE Form the context $\zeta=(I(A_{\alpha^{\ast}}),J(B_{\alpha^{\ast}}))$
\STATE Encode $T(z,x)$ using the context $\zeta$
\IF{$T(z,x)==1$}
\STATE Append to $L$ every neighbor $(z_n,x_n)$ of $(z,x)$ for which $K(z_n,x_n)=0$ and $M(z_n,x_n)=0$
\STATE If $(z_n,x_n)$ is appended, $M(z_n,x_n) \leftarrow 1$
\ENDIF

\STATE{Update R:} $R(z,x) \leftarrow T(z,x)$ 
\STATE{Update K:} $K(z,x) \leftarrow 1$ 
\STATE Remove $(z,x)$ from $L$
\ENDWHILE
\end{algorithmic}
\end{algorithm}

One of the most efficient ways of utilizing the contextual information that is needed in Stage II is described here. In order to show the elements forming the context, it is illustrated in Fig. \ref{fig:2}(c) the ground truth occupancies for the current section ($y=y_0$) and the past section ($y=y_0-1$), showing in white the (true) occupied pixels in these sections. On Fig. \ref{fig:2}(d), we show the same area during the reconstruction process, which advances section by section such that, at the moment of reconstructing the section $y=y_0$, the section $y=y_0-1$ is fully known, and the second part of the context, called matrix $B$, can be extracted and contains the true occupancy status at section $y=y_0-1$. The context $A$ from the current section for every candidate pixel $(z,x)$ is extracted from the $3\times3$ neighbourhood of the pixel (i.e., the pixels in the red square). %The status of the pixels in this reconstruction process is stored in two auxiliary binary images: the first image $K$ reveals  if the pixel  occupancy is known, and the second image $R$ stores and updates the reconstruction values at this stage.

When encoding the blue marked pixel, the pixels considered as the $3\times3$ A matrix part of context are those surrounded by the red contour. Each pixel might be green (already encoded in Stage I), for which the status is known and occupied ($K(z,x) = 1$ and $R(z,x) = 1$), khaki for forbidden pixels with status known and unoccupied ($K(z,x) = 1$ and $R(z,x) = 0$), and finally magenta  for feasible i.e., not yet known ($K(z,x) =0$). The context extracted from the current section forms the matrix A and that from the past section forms the matrix B (which are later rotated for normalizing and are combined to form the final contexts).

The procedure for encoding the points at a section $y=y_0$ with normalized contexts is summarized in Algorithm 1. Consider that we need to encode $T(z, x)$. The first part of the context uses the values of the already reconstructed pixels that are 8-neighbors of $(z, x)$ and also the information about which of the pixels were already known. The values of the pixels in the ternary image $R_k = R + K$ have the following significance: $R_k(z, x) = 0$ if the value of $T(z, x)$ is not known yet; $R_k(z, x) = 1$ if the value of $T(z, x)$ was encoded and $T(z, x) = 0$; $R_k(z, x) = 2$ if the value of $T(z, x)$ was encoded, and $T(z, x) = 1$. We consider first the 3×3 square, centered at $(z, x)$, cropped from the image $R_k$, and we denote it as a 3×3 matrix A. The elements of A belong by construction to {0, 1, 2}. The second part of the context is the 3 × 3 binary matrix B formed from P at (z, x). The information from A and B is used to form the context. For example scanning by columns we get a one-to-one correspondence between A and  $I(A)=\sum_{j=0}^{2} \sum_{i=0}^{2} A_{ij} 3^{i+3j}$. Similarly there is a one-to-one correspondence between B and $J(B)=\sum_{j=0}^{2} \sum_{i=0}^{2} B_{ij} 2^{i+3j}$. We combine them to a context label $\zeta = (I(A), J(B))$. 

The context information in A and B is further normalized in the following way: We consider performing context collapsing operations, such that if we would perform a rotation by $\alpha \in \{0, \pi/2, \pi, 3\pi/2\}$ of each of the images $R$, $T$ and $K$ around the pixel $(z,x)$, the value of the resulting normalized context is the same. We consider first the 3×3 matrix A. Apply the $\alpha$ rotation around the middle pixel and denote $A_\alpha$ the resulting 3 × 3 matrix. Compute for each of $\alpha \in \{0, \pi/2, \pi, 3\pi/2\}$, the matrix $A_\alpha$ and the weighted score of it $W(A_\alpha)$ %(favoring for example the values near the element (0, 0)) 
and pick as canonical rotation that $\alpha_\ast$ for which the weighted score $W(A_\alpha)$ is the largest. Hence, the four rotated matrices $A_\alpha$ with $\alpha \in \{0, \pi/2, \pi, 3\pi/2\}$ will be represented only by $A_{\alpha\ast}$. This process of collapsing the four matrices into a single one is performed offline once, resulting in a tabulated mapping $\alpha_\ast \leftrightarrow A$ and another mapping $I_\ast \leftrightarrow A$, which realize the mappings of the context to the canonical one, stored in look-up tables. As an example of the weighting score $W(A)$, we consider the vector $v=[A_{00} A_{01} A_{10} A_{02} A_{11} A_{20} A_{12} A_{21} A_{22} ]$  and form $W(A) = \sum_{k=0}^8 v_k 3^k$, giving in this way a larger weight to those elements which are close to the corner $(0,0)$ of A.
The normalized context is found in the following way: At each point $(z, x)$ the matrix A is formed from $R$+$K$ and the canonical rotation index $\alpha_\ast$ for this matrix is computed. %(or picked from the lookup table, at the address formed from the 9 elements of A).%
Also the corresponding rotated matrix $A_0$ is computed. The second part of the context is the 3 × 3 matrix B formed from P around $(z, x)$. The matrix B is rotated by the previously determined $\alpha_\ast$ around its center, yielding a matrix $B_0$. Now the context to be used for encoding $T(z, x)$ is constructed from $A_0$ and $B_0$ as context $\zeta = (I(A_0), J(B_0))$.

\begin{table}

\caption{Average Rate for the first 200 frames from MVUB \cite{upperbodies} and 8i \cite{8ivoxelized} datasets for the proposed Bounding Volume Lossless (BVL) encoder compared to recent codecs.}
\begin{center}

 \begin{tabular}{|c||c|c|c|c|} 
 \hline
&  \multicolumn{4}{|c|}{Average Rate [bpv]} \\\cline{2-5}
\textbf{Sequence}  & \textbf{P(PNI)\cite{garcia2019geometry}} & \textbf{TMC13\cite{gpcc}} & \textbf{DD\cite{dyadic}} & \textbf{BVL} \\
 \hline
   \multicolumn{5}{|c|}{Microsoft Voxelized Upper Bodies \cite{upperbodies}}\\\hline
Andrew9 & 1.83 & 1.14 & \textbf{1.12} & 1.17 \\ 
 \hline
 David9 & 1.77 & 1.08 & \textbf{1.06 } & 1.10\\
 \hline
 Phil9 & 1.88 & 1.18 & \textbf{1.14 } & 1.20 \\
 \hline
 Ricardo9 & 1.79 & 1.08 & \textbf{1.03} & 1.05 \\
 \hline
 Sarah9 & 1.79 & \textbf{1.07} & \textbf{1.07} & 1.08 \\ 
 \hline
\textbf{Average } & 1.81 & 1.11 & \textbf{1.08} & 1.12 \\
 \hline
  \hline
     \multicolumn{5}{|c|}{8i Voxelized Full Bodies \cite{8ivoxelized}}\\\hline
  Longdress & 1.75 & 1.03 & 0.95 & \textbf{0.91}  \\
 \hline
   Loot & 1.69 & 0.97 & 0.91 & \textbf{0.88}  \\ 
 \hline
    Redandblack & 1.84 & 1.11 &\textbf{ 1.03} & \textbf{1.03}  \\ 
 \hline
     Soldier & 1.76 & 1.04 &\textbf{ 0.96} & \textbf{ 0.96}  \\ 
 \hline
     \textbf{Average} & 1.76 & 1.04 &  0.96 & \textbf{ 0.94}  \\ 
 \hline

\end{tabular}

\end{center}
\end{table}

\subsection{Repetitive peeling-off process for complete reconstruction of more complex point clouds }
\label{sec:peeloff}

After Stage II, the reconstruction contains all the points forming the outer surfaces of the point cloud and all the inner points connected to these outer surface points, i.e., all points that are connected by a path in 3D space (in 26-voxel connectivity), to the initial points recovered in Stage I from the two depthmap images. However, there are complex point clouds, for example those representing a building and objects inside, where some objects are not connected by a 3D path to the outermost points. In that case one can repeat the encoding process shell by shell, in a peeling-off operation, where we encode first the outermost shell, defined by the points represented in the maximal and minimal depthmaps and all points connected to these points, and then we reapply the same process to the remaining un-encoded points. If needed, this peeling-off process can be applied several times. In this work, there are maximally 2 shells peeled-off and the remaining points (if any) are simply written in binary representation into the bitstream.

\section{Experimental Work}
\label{sec:expwork}

The algorithm is implemented in C and the experiments were carried out on two point cloud datasets namely, 8i Voxelized Full Bodies \cite{8ivoxelized} and Microsoft Voxelized Upper Bodies \cite{upperbodies}. The average bits per occupied voxel results (average rates) are presented on Table 1. For each point cloud, all 6 possible permutations of the 3 dimensions are tried and the best rate obtained is kept and reported here. It is observed that, proposed method performs better than the other methods on the 8i dataset. On the other hand, on MVUB dataset, our results are slightly worse than TMC13 \cite{gpcc} and Dyadic Decomposition \cite{dyadic}. Additionally, we test BVL and TMC13 on all the point clouds from the Cat1A MPEG Database \cite{ctc} having original resolutions of 10 and 11 bits. These were quantized to 10, 9 and 8 bits as well to test the performance in lower resolutions. BVL outperformed TMC13 in average at 10 bits by 6.6\%, at 9 bits by 5.2\%, at 8 bits by 2\%. At 11 bits, TMC13 outperformed BVL by 1.8\%. For all of the point clouds mentioned in this work, the decoding resulted in a perfect lossless reconstruction.  %For sparse point clouds, the performance of BVL is expected to be worse than TMC13.
%one reason being that CERV \cite{cerv} is optimized for encoding contours of piecewise constant regions and in the case of sparse point clouds the number of such regions becomes too large.

Encoding and decoding durations for BVL were measured to be 7.3 sec and 7.8 sec, respectively. On the same machine, encoding with TMC13 took 1.1 sec. All durations are measured on a single frame of the 10 bits longdress sequence by running the algorithm 10 times and taking the median. While the durations are not competitive with TMC13, it should be noted that the execution time is not yet carefully optimized.

%\vfill
%\pagebreak

\section{Conclusions}

We proposed a lossless compression method where the first stage is constructing a bounding volume for the point cloud and the following steps succeed at adding all the remaining points at a competitive bitrate, achieving state-of-the-art results for the full body datasets, and comparable results to the current GPCC standard on the upper body dynamic datasets. 
%The transformation of the lossy method “Bounding Volumes” \cite{bv} into a lossless codec is made possible through an additional primitive, achieving state of the art results for the full body datasets, and comparable results to the current GPCC standard on the upper body dynamic datasets. Further research on refining the additional coding process is under way.
\bibliographystyle{IEEEbib}
\bibliography{refs}

\begin{thebibliography}{10}

\bibitem{schwarz2018emerging}
S.~Schwarz, M.~Preda, V.~Baroncini, M.~Budagavi, P.~Cesar, P.~A. Chou, R.~A.
  Cohen, M.~Krivoku{\'c}a, S.~Lasserre, Z.~Li, et~al.,
\newblock ``Emerging {MPEG} standards for point cloud compression,''
\newblock {\em IEEE Journal on Emerging and Selected Topics in Circuits and
  Systems}, vol. 9, no. 1, pp. 133--148, 2018.

\bibitem{mpeg}
L.~Cui, R.~Mekuria, M.~Preda, and Eu.~S. Jang,
\newblock ``Point-cloud compression: Moving picture experts group's new
  standard in 2020,''
\newblock {\em IEEE Consumer Electronics Magazine}, vol. 8, no. 4, pp. 17--21,
  2019.

\bibitem{ebrahimi2016jpeg}
T.~Ebrahimi, S.~Foessel, F.~Pereira, and P.~Schelkens,
\newblock ``{JPEG Pleno}: Toward an efficient representation of visual
  reality,''
\newblock {\em Ieee Multimedia}, vol. 23, no. 4, pp. 14--20, 2016.

\bibitem{raht}
R.~L. De~Queiroz and P.~A. Chou,
\newblock ``Compression of 3d point clouds using a region-adaptive hierarchical
  transform,''
\newblock {\em IEEE Transactions on Image Processing}, vol. 25, no. 8, pp.
  3947--3956, 2016.

\bibitem{mekuria2016design}
R.~Mekuria, K.~Blom, and P.~Cesar,
\newblock ``Design, implementation, and evaluation of a point cloud codec for
  tele-immersive video,''
\newblock {\em IEEE Transactions on Circuits and Systems for Video Technology},
  vol. 27, no. 4, pp. 828--842, 2016.

\bibitem{milani2017fast}
S.~Milani,
\newblock ``Fast point cloud compression via reversible cellular automata block
  transform,''
\newblock in {\em 2017 IEEE International Conference on Image Processing
  (ICIP)}. IEEE, 2017, pp. 4013--4017.

\bibitem{garcia2018intra}
Diogo~C Garcia and Ricardo~L de~Queiroz,
\newblock ``Intra-frame context-based octree coding for point-cloud geometry,''
\newblock in {\em 2018 25th IEEE International Conference on Image Processing
  (ICIP)}. IEEE, 2018, pp. 1807--1811.

\bibitem{garcia2019geometry}
D.~C. Garcia, T.~A. Fonseca, R.~U. Ferreira, and R.~L. de~Queiroz,
\newblock ``Geometry coding for dynamic voxelized point clouds using octrees
  and multiple contexts,''
\newblock {\em IEEE Transactions on Image Processing}, vol. 29, pp. 313--322,
  2019.

\bibitem{gpcc}
``{MPEG Group} {TMC}13,'' \url{https://github.com/MPEGGroup/mpeg-pcc-tmc13},
\newblock Accessed: 2020-03-20.

\bibitem{octree}
D.~Meagher,
\newblock ``Geometric modeling using octree encoding,''
\newblock {\em Computer graphics and image processing}, vol. 19, no. 2, pp.
  129--147, 1982.

\bibitem{bdc}
R.~Ros{\'a}rio and E.~Peixoto,
\newblock ``Intra-frame compression of point cloud geometry using boolean
  decomposition,''
\newblock in {\em 2019 IEEE Visual Communications and Image Processing (VCIP)}.
  IEEE, 2019, pp. 1--4.

\bibitem{dyadic}
E.~Peixoto,
\newblock ``Intra-frame compression of point cloud geometry using dyadic
  decomposition,''
\newblock {\em IEEE Signal Processing Letters}, vol. 27, pp. 246--250, 2020.

\bibitem{bv}
I.~Tabus, E.~C. Kaya, and S.~Schwarz,
\newblock ``Successive refinement of bounding volumes for point cloud coding,''
\newblock in {\em 2020 IEEE 22nd International Workshop on Multimedia Signal
  Processing (MMSP)}. IEEE, 2020, pp. 1--6.

\bibitem{cerv}
I.~Tabus, I.~Schiopu, and J.~Astola,
\newblock ``Context coding of depth map images under the piecewise-constant
  image model representation,''
\newblock {\em IEEE Transactions on Image Processing}, vol. 22, no. 11, pp.
  4195--4210, 2013.

\bibitem{upperbodies}
C.~Loop, Q.~Cai, S.~O. Escolano, and P.~A. Chou,
\newblock ``Microsoft voxelized upper bodies - a voxelized point cloud
  dataset,''
\newblock {\em ISO/IEC JTC1/SC29 Joint WG11/WG1 (MPEG/JPEG) input document
  m38673/M72012}, 2016.

\bibitem{8ivoxelized}
E.~d’Eon, B.~Harrison, T.~Myers, and P.~A. Chou,
\newblock ``8i voxelized full bodies - a voxelized point cloud dataset,''
\newblock {\em ISO/IEC JTC1/SC29 Joint WG11/WG1 (MPEG/JPEG) input document
  WG11M40059/WG1M74006}, 2017.

\bibitem{ctc}
S.~Schwarz, G.~Martin-Cocher, D.~Flynn, and M.~Budagavi,
\newblock ``Common test conditions for point cloud compression,''
\newblock {\em Document ISO/IEC JTC1/SC29/WG11 w17766, Ljubljana, Slovenia},
  2018.

\end{thebibliography}

\end{document}